%% file: example_paper.tex
\documentclass{article}

\usepackage{microtype}
\usepackage{graphicx}
\usepackage{subfigure}
\usepackage{booktabs} 

\usepackage{hyperref}

\usepackage[accepted]{icml2023}

\usepackage{amsmath}
\usepackage{amssymb}
\usepackage{mathtools}
\usepackage{amsthm}

\usepackage[capitalize,noabbrev]{cleveref}

\theoremstyle{plain}

\theoremstyle{definition}

\theoremstyle{remark}

\usepackage{tikz}
\newcommand{\cblock}[3]{
  \hspace{-1.5mm}
  \begin{tikzpicture}
    [
    node/.style={square, minimum size=10mm, thick, line width=0pt},
    ]
    \node[fill={rgb,255:red,#1;green,#2;blue,#3}] () [] {};
  \end{tikzpicture}%
}

\usepackage[textsize=tiny]{todonotes}
\input{math_commands.tex}

\icmltitlerunning{Policy Contrastive Imitation Learning}

\begin{document}

\twocolumn[
\icmltitle{Policy Contrastive Imitation Learning}




\begin{icmlauthorlist}
\icmlauthor{Jialei Huang}{thu,sh,qz}
\icmlauthor{Zhaoheng Yin}{hk}
\icmlauthor{Yingdong Hu}{thu}
\icmlauthor{Yang Gao}{thu,sh,qz}
\end{icmlauthorlist}

\icmlaffiliation{thu}{Department of IIIS, University of Tsinghua, Beijing, China}
\icmlaffiliation{hk}{Hong Kong University of Science and Technology, Hong Kong, China}
\icmlaffiliation{sh}{Shanghai Artificial Intelligence Laboratory, Shanghai, China}
\icmlaffiliation{qz}{Shanghai Qi Zhi Institute, Shanghai, China}

\icmlcorrespondingauthor{Yang Gao}{gaoyangiiis@mail.tsinghua.edu.cn}

\icmlkeywords{Imitation Learning, ICML}

\vskip 0.3in
]



\printAffiliationsAndNotice{}  

\begin{abstract}
Adversarial imitation learning (AIL) is a popular method that has recently achieved much success. However, the performance of AIL is still unsatisfactory on the more challenging tasks. We find that one of the major reasons is due to the low quality of AIL discriminator representation. Since the AIL discriminator is trained via binary classification that does not necessarily discriminate the policy from the expert in a meaningful way, the resulting reward might not be meaningful either. We propose a new method called Policy Contrastive Imitation Learning (PCIL) to resolve this issue. PCIL learns a contrastive representation space by anchoring on different policies and generates a smooth cosine-similarity-based reward. Our proposed representation learning objective can be viewed as a stronger version of the AIL objective and provide a more meaningful comparison between the agent and the policy. From a theoretical perspective, we show the validity of our method using the apprenticeship learning framework. Furthermore, our empirical evaluation on the DeepMind Control suite demonstrates that PCIL can achieve state-of-the-art performance. Finally, qualitative results suggest that PCIL builds a smoother and more meaningful representation space for imitation learning.
\end{abstract}

\input{introduction}
\input{background_and_notation}

\input{method}
\input{experiments}
\input{related_work}

\input{conclousion}

\nocite{langley00}

\bibliography{example_paper}
\bibliographystyle{icml2023}


\input{appendix.tex}

\end{document}

%% file: math_commands.tex
\usepackage{amsmath,amsfonts,bm}

\def\eqref#1{equation~\ref{#1}}

\def\1{\bm{1}}

\def\rva{{\mathbf{a}}}

\def\rvs{{\mathbf{s}}}

\def\rvx{{\mathbf{x}}}

\DeclareMathAlphabet{\mathsfit}{\encodingdefault}{\sfdefault}{m}{sl}
\SetMathAlphabet{\mathsfit}{bold}{\encodingdefault}{\sfdefault}{bx}{n}

\def\gA{{\mathcal{A}}}

\def\gS{{\mathcal{S}}}

\newcommand{\E}{\mathbb{E}}



%% file: introduction.tex
\section{Introduction}
Imitation is one of the fundamental capabilities of an intelligent agent~\citep{hussein2017imitation}. Animals and humans can acquire many skills by mimicking each other~\citep{animal_imitate}. In engineering, imitation learning also enables many robotics applications. One mainstream class of imitation learning algorithms is the adversarial imitation learning~(AIL)~\citep{ho2016generative}. AIL converts the imitation task into a distribution matching problem and proposes to imitate it by training a policy against an adversarial discriminator. AIL has enjoyed great success on many imitation tasks: it achieves superior performance~\citep{ho2016generative, kostrikov2018discriminator}, and has been experimentally proven to alleviate some of the distributional drift issue, and can work even without expert actions~\citep{torabi2018generative}.However, AIL is hard to train in practice, usually involving careful tuning of discriminator neural network sizes and learning rates~\citep{wang2017robust, kim2018imitation, what_matters_ail}. The fragility of the discriminator \citep{peng2018variational} not only leads to poor performance but also severely limits the applicability of AIL to a broader range of tasks. 

Numerous techniques have been proposed to improve the performance of AIL, such as using regularization and gradient penalties~\citep{fu2017learning, kostrikov2018discriminator,gulrajani2017improved}. Some works also propose to use different distribution metrics (e.g., KL divergence, Wasserstein distance)~\citep{wail} for distribution matching and show some improvements. Though these methods show encouraging results, we notice that they ignore one crucial aspect of the problem: the representation of AIL's discriminator. To be specific, the discriminator in AIL is usually trained with binary classification loss that distinguishes expert transitions from agent transitions. This discriminator is then used to define rewards. However, since the only goal of the discriminator is to distinguish the expert from the agent, it does not necessarily learn a good, smooth representation space that can provide a reasonable comparison between the behavior of two agents.Ideal representations should be able to provide semantically meaningful signals to compare the expert policy and the agent policy.


\begin{figure*}[t]
\begin{center}
\includegraphics[width=0.80\textwidth]{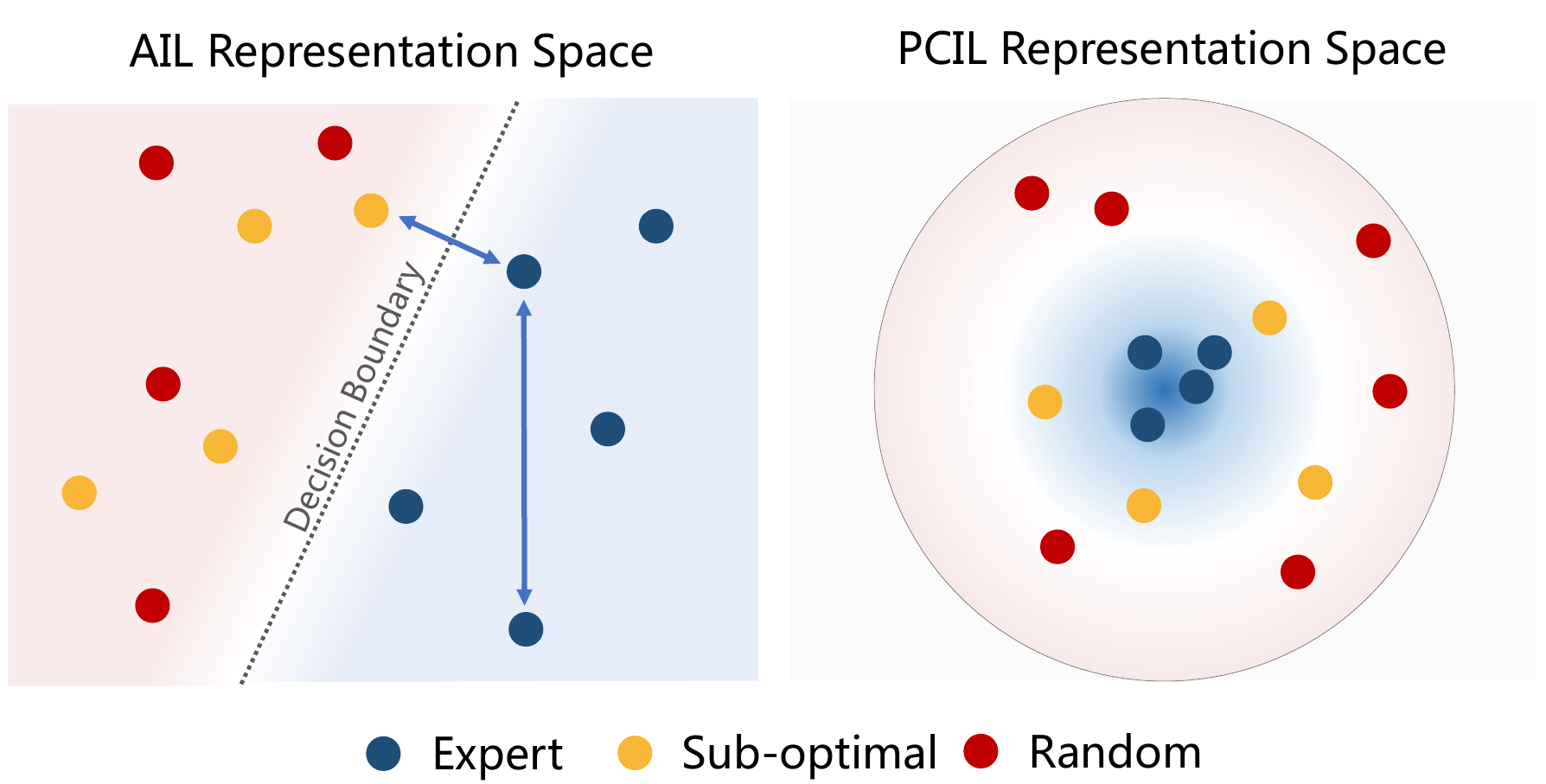}
\label{fig:teaser}
\caption{Comparison between the representation space of AIL method and our method. Since the AIL methods use a binary classification objective to distinguish expert and non-expert transitions, the representation space is only required to separate two classes in two disjoint subspaces. So the embedding space is not required to be semantically meaningful enough, e.g. \textbf{(Left)} the distance between 2 expert data points may be even longer than the distance between expert data point and sub-optimal non-expert data point. \textbf{(Right)} We overcome this limitation by proposing PCIL. Our method enforces the compactness of the expert's representation. This ensures that the learned representation can capture common, robust features of the expert's transitions, which leads to a more meaningful representation space.}
\end{center}
\end{figure*}

In this paper, we propose a new algorithm called Policy Contrastive Imitation Learning~(PCIL) to achieve this goal. Instead of training with a binary-classification objective, we propose to train a discriminator representation space with the contrastive learning loss. Our method differs from the prior representation learning approach in AIL in that we perform contrastive learning between different policies. More specifically, we push the expert's representation together and pull the agent policy's representation away from them. We define the imitation learning reward via cosine similarity between the policy's and expert's transition. 

As is shown in Figure~\ref{fig:teaser}, the discriminator~(binary classifier) might not have a good representation space: the distance between two expert transitions can be even larger than the distance between an expert transition and an agent transition. This implies that the discriminator may not encode some common features of the expert's behavior and may use non-robust features to compare the behavior. The lack of proper structure in the representation space of traditional discriminators may yield low-quality AIL rewards. To alleviate this issue, we explicitly define a constraint on the representation space, requiring that the distance between expert transitions be smaller than their distance to the agent's transitions. This is a stronger constraint on the discriminator's representation: it is easy to derive a binary classifier from our learned representation. However, the binary classifier's representation space does not necessarily satisfy our objective.

We validate PCIL by first showing that PCIL is performing distribution matching between the expert and the agent policy.  We also carry out empirical evaluation by benchmarking our method on the DeepMind Control Suite~\citep{tassa2018deepmind}. Experimental results show that our method is able to achieve state-of-the-art results. Through ablation study and qualitative visualization, we find that our method is more effective than prior representation learning methods and able to provide a better representation space for imitation learning.


In summary, our contributions in this paper are as follows. (1). We point out a new direction to improve the performance of the AIL methods, i.e., going beyond naive binary classification and leveraging more stable and meaningful representation learning algorithms for imitation. (2). We propose an algorithm called Policy Contrast Imitation Learning (PCIL) method that instantiates such an improvement and establishes its connection to apprenticeship learning from a theoretical perspective. (3).  We evaluate our method on the DeepMind Control Suite and achieve state-of-the-art performance. Through ablation studies, we highlight its essential difference from previous contrastive learning methods in AIL.

%% file: background_and_notation.tex
\section{Preliminaries}
\subsection{Notations}
In this paper, we model the imitation leanring problem as a markov decision process $\mathcal{M} = \left(\mathcal{S}, \mathcal{A}, p_0(s), p\left(s^{\prime} \mid s, a\right), r\left(s, a, s^{\prime}\right), \gamma\right)$. Here, $\mathcal{S}$ is the state space. $\mathcal{A}$ is the action space. $p_0(s)$ defines the initial state distribution. $p(s'|s,a)$ defines the transition dynamics. $r(s, a, s')$ is the reward function. $\gamma$ is the discount factor. The goal is to maximize the expected return of the learned policy $\pi$, which is defined by 
\begin{align*}
\mathcal{J}(\pi) = \mathbb{E}_{\substack{s_0\sim p_0(s), \\
a_i\sim \pi(\cdot|s_i),\\
s_{i+1}\sim p(\cdot|s_i, a_i)}}\left[\sum_{k=0}^{\infty} \gamma^{k} r\left(s_k, a_k, s_{k+1}\right)\right].
\end{align*}
For the imitation learning problem, the algorithm does not have access to the reward function and the transition dynamics. Instead, it is provided with an expert demonstration dataset $\mathcal{D}$ sampled from an expert policy $\pi_E$, which can perform well in $\mathcal{M}$. Here, $\mathcal{D}$ takes the form of $\{(s_i^E, a_i^E)\}$, where $(s_i^E, a_i^E)$ is sampled from $\rho_{E}$, the stationary state-action visiting distribution of $\pi_E$. The imitation learning algorithm is then required to reproduce the expert's behavior with $\mathcal{D}$.

\subsection{Adversarial Imitation Learning}
One popular class of the imitation learning algorithm is the adversarial imitation learning~(AIL), whose vanilla version is Generative Adversarial Imitation Learning~(GAIL)~\citep{ho2016generative}. The idea of GAIL is to minimize the divergence between $\rho_{E}$ and $\rho_\pi$. It uses a discriminator $D(s, a)$ to distinguish expert's transitions $(s^E_i,a^E_i)\sim \mathcal{D}$ from the policy transitions $(s, a) \sim \rho_\pi$, which is trained by maximizing the objective 
\begin{align*}
\mathcal{L} =\ & \mathbb{E}_{(s_i, a_i)\sim\pi}[\log (D(s_i, a_i))]+ \\
&\mathbb{E}_{(s_i^E, a_i^E)\sim \mathcal{D}}[\log (1-D(s_i^E, a_i^E))].
\end{align*}
To achieve imitation, the agent policy is then required to fool the discriminator, which can only be possible when the policy $\pi$ resembles the expert $\pi_E$. Specifically, GAIL defines an adversarial reward $r(s_t, a_t) = -\log (1 - D(s_t, a_t))$, and trains $\pi$ to maximize the expected return with respect to this reward using on-policy RL algorithms.

%% file: method.tex
\section{Policy Contrastive Imitaiton Learning}
\subsection{Overview}
\begin{figure*}[t]
\begin{center}
\includegraphics[width=0.94\textwidth]{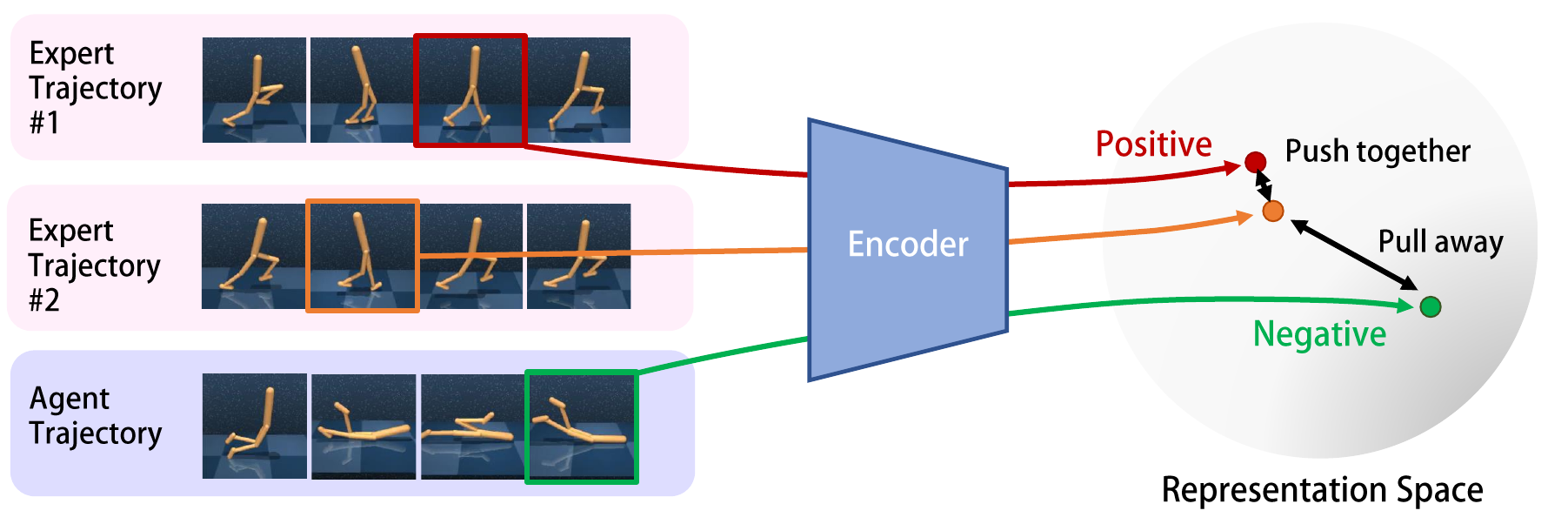}

\caption{Illustration of our contrastive learning approach. We first select an anchor state~(the orange) from the expert trajectory. Then, we select a positive state sample~(the red) from another expert trajectory and a negative state sample~(the green) from the agent trajectory. We map these selected states to the representation space. Finally, we push the representation of the anchor state and the positive state together and pull the representation of negative samples away from the representation of the anchor state.}
\label{fig:main}
\end{center}
\end{figure*}

We propose a novel representation-learning-based approach called Policy Contrastive Imitation Learning to improve the AIL reward. The overview of PCIL is illustrated in Figure \ref{fig:main}. Our key insight is to learn a policy-contrastive representation space. Unlike the contrastive learning studied in previous AIL literature, the policy-contrastive representation here is learned by anchoring on the rollout of different policies, based on which we can compare the behavior of different policies in a meaningful way. We will discuss the training of this representation in Section~\ref{subsec: SCL in AIL} and the reward design in Section \ref{subsec: cos based reward}. Then, we will show the convergence of our algorithm in Section \ref{sec:theory}. 

\subsection{Contrastive Policy Representation for Imitation}
\label{subsec: SCL in AIL}
The vanilla AIL algorithms are based on unconstrained representations and can be very non-robust. One possible approach to handle this problem is to learn a more meaningful representation by contrastive learning. Researchers~\citep{he2020momentum} have found that it can usually learn semantically meaningful representation, leading to better performance on downstream tasks, such as classification. However, though it is effective in the field of supervised learning, prior work~\citep{rep_study} has found that it does not greatly improve AIL much.

To understand the reason behind this, we first recall that the contrastive learning method learns by drawing the representation of one training sample $\rvx$ towards a similar positive sample $\rvx_p$, and pulling it away from a dissimilar, negative sample $\rvx_n$. Some heuristic rules determine the choice of positive and negative samples: the positive sample is usually defined as a data augmentation of $\rvx$. Therefore, this augmentation decides what should be similar to the representation. Nevertheless, we argue that the definition of similarity in the previous works is not strong enough for AIL. This is because, in AIL, the representation should also be able to let us discern good behavior from the bad behavior. However, the difference between good and bad behavior here can sometimes be very faint. For example, consider a case where a robot misses the exact point to execute a certain action and, as a result, fails to accomplish the task. The good and bad states right before this point may look very similar. More specifically, the difference can simply be minor in a particular physical measurement, like the distance. In this case, the representation should consider this as a semantic component and be sensitive to such a difference to succeed. Unfortunately, the difference between the positive sample and the anchor sample in the previous representation learning methods is usually very large and overwhelms the difference between good and bad states. As a result, the model may not distinguish between good and bad states effectively. Though combining the representation learning objective and the AIL objective may help combat this problem by enforcing a hard distinguishing constraint over contrastive representation, in practice we find this does not work well~(Section~\ref{subsec: ablation}). 

These observations motivate us to learn a representation that is semantically meaningful and able to distinguish between good and bad states. We find a surprisingly simple yet very effective approach: we can consider the samples drawing from the same policy as the positive samples and the samples from all the other policies as negative samples. In the case of imitation learning, our samples can be naturally divided into two categories, namely expert and non-expert samples. Then given the encoder $\Phi:\mathcal{S}\to\mathbb{S}$ that maps the state to a representation vector in a high-dimensional sphere, we define its infoNCE representation loss function as follows:

\begin{equation}
\begin{aligned}
    &\mathcal{L} = \mathbb{\E}_{\substack{\rvx_0=(\rvs, \rva)\sim\mathcal{D}, \\ \rvx_p=(\rvs_p, \rva_p)\sim\mathcal{D},\\ \tilde{\rvx}_i=(\tilde{\rvs}_i, \tilde{\rva}_i)\sim\rho_{\pi}}} \left[ \Phi(\rvx_0)^T\Phi(\rvx_p) + \right.  \\
    &\left.{\log \left(\exp \Phi(\rvx_0)^T\Phi(\rvx_p) + \sum_{i=1}^n \exp \Phi(\rvx_0)^T\Phi(\tilde{\rvx}_i)\right)} \right]. 
\end{aligned}
\label{eq:loss}
\end{equation}

Here, $\rvx_0$ is some state-action pair from the expert transitions; $\rvx_p$ is some transition from the expert data acting as positives; $\tilde{\rvx}_i$ is some agent transition working as negatives. In other words, we require $\Phi$ to draw the expert samples towards each other and pull all the policy samples away from the expert samples. 

Our proposed objective function is a strictly stronger constraint on the discriminator. The binary classification discriminator can find any hyper-plane that separates the two types of transitions, with no constraint on how the transitions are embedded. However, our objective enforces the pair-wise distance constraint between any triplets. One can derive a binary expert-policy transition classifier from a trained $\Phi$ by computing $ \Phi(\rvx_0)^T \Phi(\rvx) > t$, where $\rvx_0$ is any expert transition, $\rvx$ is the transition to be classified and $t$ is some threshold. However, on the other hand, the binary classification induced latent space might not satisfy our constraint. As illustrated in Figure~\ref{fig:teaser}, the latent space for the binary classification discriminator might have some expert-expert pairs that are even further away than some expert-agent pairs.  


Interestingly, our approach echoes the supervised contrastive learning (SCL)~\citep{khosla2020supervised}, which suggests that we consider all the samples in one class as similar, positive samples. Here we also consider all the samples in the expert demonstrations as similar. However, unlike SCL, we do not require the samples from the agents to be similar to each other. This is because the agent transitions are generated from different policies during the training process.
\vspace{-0.6cm}
\subsection{Similarity-Based Imitation Reward}
\label{subsec: cos based reward}
With a representation that can capture the difference between good and bad states, we can then define a reward function to encourage imitation learning. Though using an AIL-style reward with this representation is still possible, we find that a better choice is to use a cosine similarity metric to define the reward. It has several advantages: it is bounded and appears relatively smooth in practice, leading to more stable learning. Concretely, we define:
\begin{equation}
    r(\rvx) =  \Phi\left(\rvx\right)^T \mathbb{E}_{\rvx_E\sim\mathcal{D}}\Phi\left(\rvx_E\right) . 
    \label{eq:rew}
\end{equation}
Nevertheless, in practice evaluating the latter expectation can be time-consuming since $\Phi$ is frequently updated. Therefore, we use a random expert sample for the reward calculation. From this reward, we can see that a policy can only obtain high rewards when it frequently visits the expert's distribution. This naturally connects our method to the distribution matching, and we provide the theoretical analysis of our algorithm in the following subsection.

\subsection{Theoretical Analysis}
\label{sec:theory}

In this part, we first show that PCIL can be reduced to Apprenticeship Learning~(AL)~\citep{abbeel2004apprenticeship}. Based on this observation, we can further show that PCIL is also perform distribution matching by minimizing a divergence $\rho^E$ and $\rho^\pi$ that is equivalent to the total variation divergence. 

Recall that an AL problem takes the following form~\citep{ho2016generative}: 
\begin{equation}
\underset{\pi}{\min} \max _{r \in \mathcal{R}} \displaystyle \mathop{\E}_{\substack{\rvx=(\rvs,\rva)\sim\mathcal{D}}}\left[r(\rvx)\right] - \displaystyle \mathop{\E}_{\substack{\rvx=(\rvs,\rva)\sim\rho_{\pi}}}\left[r(\rvx)\right],
\label{eqn:target}
\end{equation}
where $\mathcal{R}$ is a set of reward functions. AL plays a min-max game between the policy $\pi$ and the reward function $r$. Intuitively, in the inner loop we would like to find a cost function such that the expert data's cummulative return is higher than that of the agent's, and their gap is maximized. Meanwhile, the policy $\pi$ tries to minimize this gap. 

Now, we reduce our objective to the AL formulation. For simplicity, we consider the case that we only have one negative sample. We notice that Equation~\ref{eq:loss} is then
\begin{equation}
    \mathcal{L} = \displaystyle \mathop{\E} \left[ - \log \frac{\exp \Phi(\rvx_0)^T\Phi(\rvx_p)}{ \left(\exp \Phi(\rvx_0)^T\Phi(\rvx_p) + \exp \Phi(\rvx_0)^T\Phi({\rvx}_n\right)} \right ]
\end{equation}
As suggested by \citep{khosla2020supervised}, we can apply the Taylor expansion trick to approximate this loss function with the following form:
\begin{equation}
\mathcal{L} \approx  \mathbb{E}~ [\Vert\Phi(\rvx_0) - \Phi(\rvx_p)\Vert^2 - \Vert \Phi(\rvx_0) - \Phi({\rvx}_n)\Vert^2 ].
\label{eq:pf3}
\end{equation}  
Note that we drop the constant terms and the scaling constant since they do not affect the optimization objective. Moreover, since $\Phi$ embeds the data points to the sphere, we have $\Vert\Phi(\rvx)\Vert^2=1, \forall \rvx$. As a result, we can further expand each term above and have
\begin{equation}
\mathcal{L} = \mathbb{E} \left[\Phi(\rvx_0)^T\Phi(\rvx_n) - \Phi(\rvx_0)^T\Phi({\rvx}_p)\right].
\label{eq:pf4}
\end{equation}
Since the variables in this equation are independent from each other, we are minimizing
\begin{equation}
    \begin{split}
    \mathcal{L} = \mathbb{E}_{\rvx_n\sim\rho_\pi} [\mathbb{E}_{\rvx_0\sim\mathcal{D}}[\Phi(\rvx_0)]^T\Phi(\rvx_n)] -\\
    \mathbb{E}_{\rvx_p\sim\mathcal{D}} [\mathbb{E}_{\rvx_0\sim\mathcal{D}}[\Phi(\rvx_0)]^T\Phi({\rvx}_p)].
    \end{split}
    \label{eq:fin}
\end{equation}
Let the reward function $r_\theta(\rvx) = \mathbb{E}_{\rvx_0\sim\mathcal{D}}[\Phi_\theta(\rvx_0)]^T\Phi_\theta(\rvx)$ as we defined in Equation~\ref{eq:rew}, then minimizing the Equation~\ref{eq:fin} is exactly doing the maximization of 
\begin{equation}
\displaystyle \mathbb{E}_{\rvx\sim\mathcal{D}}\left[r_\theta(\rvx)\right] - \displaystyle \mathbb{E}_{\rvx\sim\rho_{\pi}}\left[r_\theta(\rvx)\right],
\end{equation}
which is exactly the inner maximization loop of AL. Then optimizing the policy with respect to this $r_\theta$ is exactly the outer loop. Hence, our algorithm is reduced to AL. 

Finally, we can show that we are minimizing the total variation divergence between the expert and policy distribution as follows. First, let us define the inner objective of Equation \ref{eqn:target} as
$$
D_{cont}(\rho^E||\rho^\pi) = \max_{r_\theta} [\mathbb{E}_{\rvx\sim \rho^E} r_\theta(\rvx) - \mathbb{E}_{\rvx\sim \rho^\pi} r_\theta(\rvx)].
$$
Here we assume that the expert dataset $\mathcal{D}$ is large enough, so sampling from $\mathcal{D}$ can be regarded as sampling from $\rho^E$. Then we can show that \ \\ \\
\textbf{Theorem 1} (Equivalence to TV divergence)
$$
    0.25 D_{TV}(\rho^E||\rho^\pi)  \leq D_{cont}(\rho^E||\rho^\pi) \leq 2.0 D_{TV}(\rho^E||\rho^\pi).
$$
Here, $D_{TV}(\rho^E||\rho^\pi) = \frac{1}{2}\int_\mathcal{X} |\rho^E(x) - \rho^\pi(x)|dx$ is the total variation divergence. Therefore, our objective is a proxy of $D_{TV}(\rho^E||\rho^\pi)$.  \\

We leave the proof of the theorem to Appendix~\ref{ap:proof}.

%% file: experiments.tex
\section{Experiments}
In this section, we empirically evaluate PCIL on an extensive set of tasks from the DeepMind control suite~\citep{tassa2018deepmind}, a widely used benchmark for continuous control. Our experiments are designed to answer the following questions: (1) Can PCIL achieve expert performance, and how sample efficient is PCIL compared to state-of-the-art imitation learning algorithms? (2) How does the representation space of PCIL differ from that of the AIL methods? (3) How does our method perform when we use different representation learning methods and reward design?

\subsection{Experimental Setup}
\label{subsec: setup}

\paragraph{Environments}
We experiment with 10 MuJoCo~\citep{todorov2012mujoco} tasks provided by DeepMind Control Suite. The selected tasks cover various difficulty levels, ranging from simple control problems, such as the single degree of freedom cart pole, to complex high-dimensional tasks, such as the quadruped run. The episode length for all tasks is 1000 steps, where a per-step ground truth environment reward is in the unit interval $[0, 1]$. 
For each task, we train an expert policy using DrQ-v2~\citep{yarats2021mastering} with the true environment reward function and use it to collect 10 demonstrations. We refer readers to Appendix~\ref{sec:env} for the full task list and more details about the demonstrations.

\paragraph{Training Details}
To update the encoder, we randomly sample 128 expert transitions and 128 agent transitions from a replay buffer. For arbitrary expert transition, any other expert transition is considered a positive sample, and all the agent transitions constitute the set of negative samples. We update the encoder by minimizing Equation~\ref{eq:loss} with respect to these samples. We use DrQ-v2~\citep{yarats2021mastering} as the underlying RL algorithm to train the agent with the cosine similarity reward given in Equation~\ref{eq:rew}. We use a budget of 2M environment steps for all the experiments. Further implementation details can be found in Appendix~\ref{sec:algodetails}.




\paragraph{Baselines}
We compare PCIL to Behavioral Cloning (BC) and two major classes of imitation learning algorithms:

\begin{enumerate}
    \item \textbf{Adversarial IRL:}
    We consider Discriminator-Actor-Critic (DAC)~\citep{kostrikov2018discriminator}, a state-of-the-art AIL method that employs an unbiased AIL reward function and performs off-policy training to reduce environmental interactions.
    
    \item \textbf{Trajectory-matching IRL:} Primal Wasserstein Imitation Learning (PWIL)~\citep{dadashi2020primal} and Sinkhorn Imitation Learning (SIL)~\citep{papagiannis2020imitation} are two recently proposed trajectory-matching imitation learning methods. PWIL computes the reward based on an upper bound of Wasserstein distance. SIL computes the reward based on Sinkhorn distances~\citep{cuturi2013sinkhorn}.
\end{enumerate}
To ensure a fair comparison, we implement all the baselines using the same RL algorithm. The implementation details of these algorithms are in the Appendix.

\begin{figure*}[t]
\begin{center}
\includegraphics[width=\textwidth]{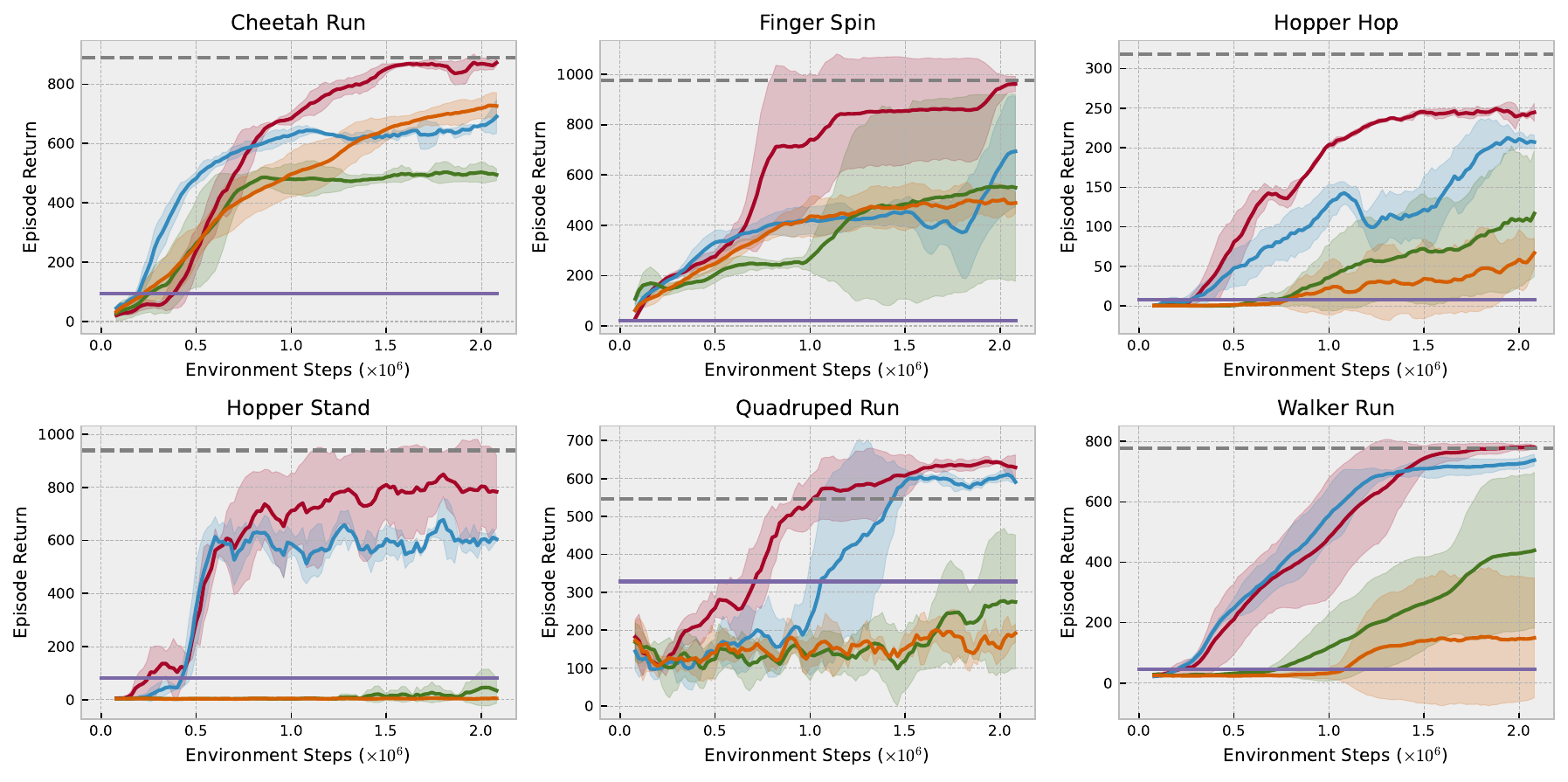}
\cblock{128}{128}{128}\hspace{1mm}Expert\hspace{1.5mm}
\cblock{122}{104}{166}\hspace{1mm}BC\hspace{1.5mm}
\cblock{213}{94}{0}\hspace{1mm}SIL\hspace{1.5mm}
\cblock{70}{120}{33}\hspace{1mm}PWIL\hspace{1.5mm}
\cblock{52}{138}{189}\hspace{1mm}DAC\hspace{1.5mm}
\cblock{166}{6}{40}\hspace{1mm}PCIL (Ours)\hspace{1.5mm}
\end{center}
\caption{Comparisons of algorithms on 6 selected tasks. See Appendix~\ref{sec:more-results} for more tasks. For every 20k environment steps, we perform 10-episode rollouts of the policy without exploration noise and report average episode returns over the 10 episodes. We plot the mean performance over 3 seeds together with the shaded regions, which represent 95\% confidence intervals.}
\label{fig:bench}
\end{figure*}

\subsection{Main Results}
We show the performance curves of 6 tasks in Figure~\ref{fig:bench}, which are averaged over three random seeds. More results on DeepMind control tasks are provided in Appendix~\ref{sec:more-results}. We find that PCIL is able to outperform the existing methods on all of these tasks. It achieves near-expert performance within our online sample budget in all considered tasks except Hopper Hop. In terms of sample efficiency, i.e., the number of environment interactions required to solve a task, PCIL shows significant improvements over prior methods on five tasks: Cheetah Run, Finger Spin, Hopper Hop, Hopper Stand, and Quadruped Run. For the remaining tasks, PCIL achieves similar results compared with the state-of-the-art adversarial imitation learning method DAC. In particular, we notice that PCIL's performance gain is larger on more difficult tasks (e.g., Cheetah Run, Quadruped Run). On those easier tasks (e.g., Walker Stand, Walker Walk), the baselines are also able to achieve strong results.

\subsection{Analysis of Representation Space}

\begin{figure*}[tb]
\centering
\includegraphics[width=\textwidth]{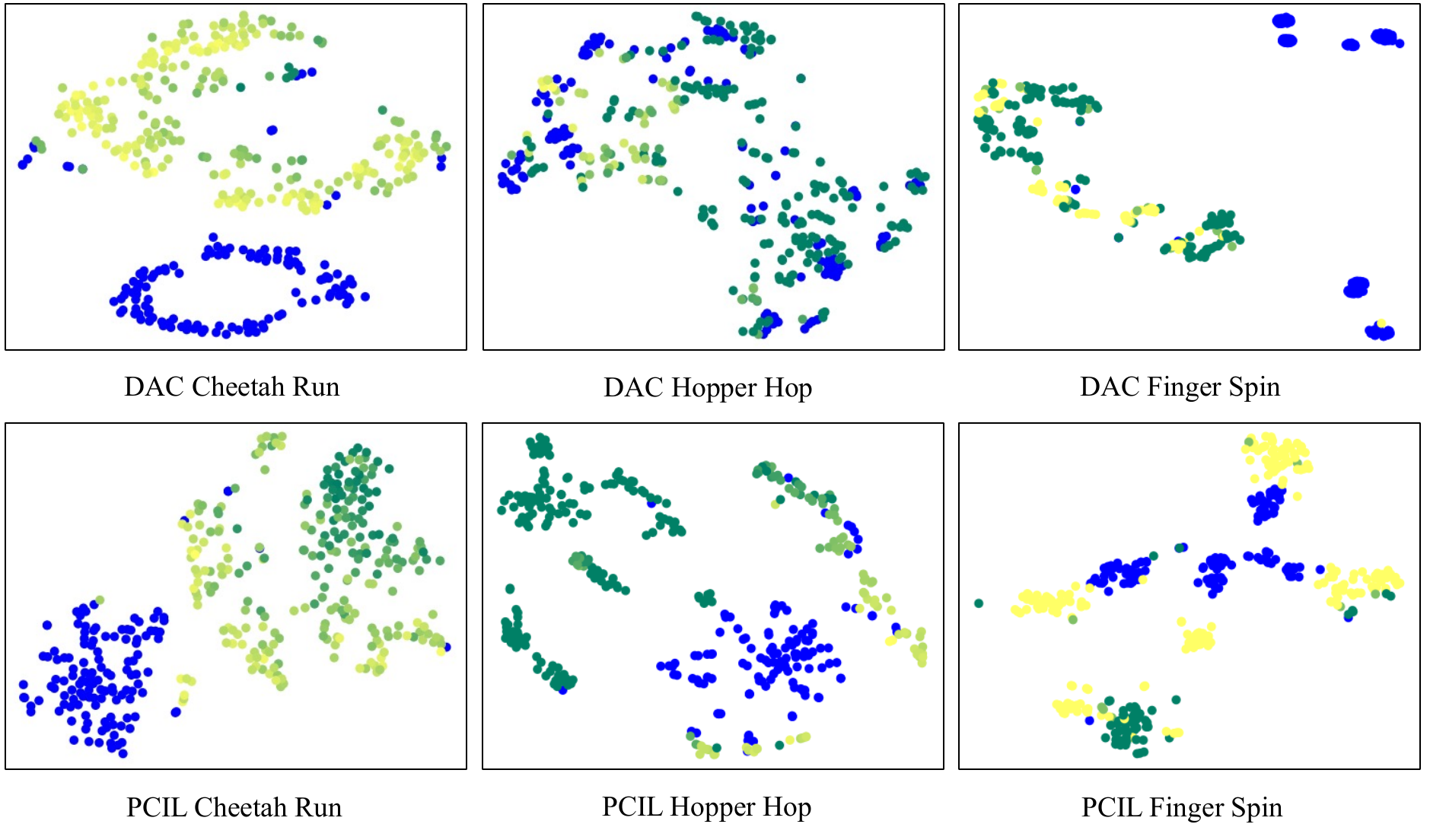}
\caption{t-SNE visualization results for DAC (top panels) and our PCIL method (bottom panels). The blue color indicates the expert's transition. The lighter~(yellow) color indicates agent transitions with higher real reward while the darker~(green) indicates lower real reward.}
\label{fig:visualize}
\end{figure*}

We visualize the representation space of PCIL and DAC using 
t-SNE~\citep{van2008visualizing}
in Figure~\ref{fig:visualize}. For DAC, since there is no explicit representation learning in the discriminator, we treat the last hidden layer of its discriminator as the representation. We randomly sample 128 expert transitions and 256 agent transitions for visualization. For a fair comparison, all the transitions of PCIL and DAC are selected from the same episode during the training process on the same task. We use color to indicate the real environment reward of the agent's transitions. The lighter~(yellow) color indicates a higher reward for agent transition while the darker~(green) indicates a lower reward agent transition. The blue color indicates the expert's transition.

We observe that in the representation space of PCIL, the expert transitions are concentrated in a cluster. Moreover, the distance between each agent transition to the cluster of expert transition is highly correlated with the real reward of that agent transition. This fact suggests that the contrastive objective of PCIL indeed induces a meaningful representation space here. 
On the contrary, the representation space of DAC is much less structured. The expert transitions are scattered throughout the representation space of DAC. Moreover, we identify that in the DAC's representation space, the agent's real ward does not correlate well with its distance to the expert transitions. These facts show that our method indeed learns a better representation space.




\subsection{Ablation Studies}
\label{subsec: ablation}
As described in our method, our method has two components: a policy contrastive representation for imitation (Section~\ref{subsec: SCL in AIL}) and a similarity-based imitation reward (Section~\ref{subsec: cos based reward}). In this part, we carry out ablation studies to analyze their effects. We first introduce our design choices as follows. 

\begin{table*}[t]
\caption{Ablation studies on the policy contrastive representation and similarity-based imitation reward. We report the average final returns on 4 selected tasks over 3 random seeds and standard deviations are given in error bars.}
\vskip 0.15in
\centering
\begin{tabular}{lcccc}
\toprule
\bf Methods & \bf Finger Spin & \bf Walker Run & \bf Hopper Stand & \bf Hopper Hop\\
\midrule
PCL + Sim-reward (Ours)                 & $\bf964.2 \pm 24.2$ & $\bf778.4 \pm 14.1$  & $\bf771.2 \pm 142.1$ & $\bf43.2 \pm 20.1$ \\
TCN + Sim-reward                        & $\bf 975.2 \pm 45.1$ & $180.5 \pm 11.4$ & $2.0 \pm 1.2$ & $6.8 \pm 0.2$ \\
PCL + GAIL-reward                       &$2.5 \pm 2.4$    & $18.5 \pm 6.5$  & $1.2 \pm 1.8$ & $0.1 \pm 0.1$ \\
TCN + GAIL-reward                       &$35.4 \pm 19.2$  & $19.8 \pm 4.2$ & $1.5 \pm 0.3$ & $17.3 \pm 4.6$\\
\bottomrule
\end{tabular}
\vskip -0.1in
\label{table:ablation}
\end{table*}

\paragraph{PCIL Representation v.s. TCN Representation}
We replace our proposed policy contrastive objective with other contrastive learning methods. For this purpose, we adopt the popular self-supervised representation learning method that leverages temporal information: Time-Contrastive Networks (TCN)~\citep{sermanet2018time}. In this case, the positive samples are selected within a small window around the anchor sample, while the negative samples are selected from distant time steps in the same rollout trajectory. See Appendix~\ref{subsec: ablation detail} for implementation details.

\paragraph{Similarity-based Reward v.s. GAIL-like Reward}
We also ablate the similarity-based reward in PCIL by replacing the similarity-based imitation reward with a GAIL-like reward. Specifically, we train a linear binary classifier on the policy contrastive embedding space to distinguish expert or non-expert data. In this case, the embedding space is still trained by PCIL contrastive loss, and the GAIL reward's gradient is detached from the embedding network. We use the same reward predictor as other GAIL-style methods~\citep{kostrikov2018discriminator}, i.e. $\log(D(x))-\log (1-D(x))$. 

\paragraph{Analysis} By comparing rows 1 and 2 in Table~\ref{table:ablation}, we find that the approach with TCN encoder does not work in three out of four environments. This is because the optimization goal of TCN is not to distinguish between expert and non-expert data. Thus the reward produced by comparing expert and non-expert data in the learned representation space is not necessarily meaningful. Note that the case in row 2 is no longer an adversarial IRL method. We also consider a case (row 4 in Table~\ref{table:ablation}) where we use TCN and GAIL-like reward predictor, but the performance of this method is poor.
Moreover, we observe that in the absence of the similarity-based imitation reward (compare rows 1 and 3 in Table~\ref{table:ablation}), our method does not work. This is because our representation space has metric-space characteristics. As a result, we should use a distance-based metric to compute the reward. In conclusion, the two components of our method are necessary for achieving good performance.

%% file: related_work.tex
\section{Related Work}


\subsection{Imitation Learning}
Imitation learning is a class of algorithms that enables a robot to acquire behaviors from a demonstration dataset. There are two classes of imitation learning algorithms: behavioral cloning~(BC) and Inverse Reinforcement Learning~(IRL)~\citep{ng2000algorithms}. BC is a simple supervised learning algorithm that directly fits the expert's action. However, some work suggests that it has some drawbacks: it suffers from covariate shift problem~\citep{ross2011reduction}, and it is hard to learn from a demonstration dataset without expert actions~\citep{torabi2018behavioral}. Instead, IRL \citep{abbeel2004apprenticeship} proposes to recover the underlying policy by estimating the underlying reward function and then maximizing the overall return with this reward. In particular, a recent branch of IRL is the AIL, which proposes to match agents' state-action distribution with experts via adversarial training. GAIL~\citep{ho2016generative} proposed a maximum entropy occupancy measure matching method which learns a discriminator to bypass the need to recover the expert's reward function. Later, several works proposed an improved version of the GAIL methods~\citep{ghasemipour2020divergence, blonde2022lipschitzness, baram2017end, kostrikov2018discriminator,fu2017learning}. AIRL~\citep{fu2017learning} replaced the Shannon-Jensen divergence used in GAIL by KL divergence to measure similarity between state-action pair distributions. \cite{baram2017end} bridges the GAIL framework to model-based reinforcement learning. DAC~\citep{kostrikov2018discriminator} improved the sample efficiency by leveraging a replay buffer without importance sampling and dealing with the absorbing state problem. In contrast to these works, we focus on the representation of AIL's discriminator and reformulating AIL in a contrastive embedding space.

\subsection{Representation Learning For Policy Learning}
In this work, we propose a representation-learning-based approach to improve imitation learning. As imitation learning is a major class of policy learning algorithms, we review works that use self-supervised learning to improve policy learning in this part.
Pioneer works~\citep{mirowski2016learning, jaderberg2016reinforcement, shelhamer2016loss, lample2017playing} explore using auxiliary objectives (e.g., predict some property of the environment) as a self-supervision signal.
Recent works employ more general self-supervised objectives~\citep{oord2018representation}. In particular, \cite{srinivas2020curl} are based on contrastive representation learning. \cite{sermanet2018time} learn representation from multiview video using time contrastive learning.
Some other methods also explore the use of self-supervised representation pretrained on environment data~\citep{ha2018world} or from real-world images~\citep{xiao2022masked, parisi2022unsurprising, nair2022r3m}.
In imitation learning, \cite{mandi2021} proposes to use contrastive learning for one-shot imitation learning in robotics. \cite{rep_study} investigate the use of representation for imitation learning. However, their result suggests that self-supervised representation learning only provides a small improvement of imitation learning algorithms' performance. Our method differs from all these existing works by proposing to anchor on different policies and learn a discriminative self-supervised representation for imitation learning. 


%% file: conclousion.tex
\section{Conclusion}
In this paper, we suggested a new approach to improve adversarial imitation learning algorithms: to learn a more meaningful, discriminative representation space for imitation. To this end, we proposed a new algorithm called PCIL. We conducted a theoretical analysis of our method and showed its connection to apprenticeship learning. We also conducted experiments on the DeepMind Control Suite and showed that PCIL could achieve state-of-the-art performance. Moreover, we used an ablation study to highlight its difference from the previous representation learning method. In the future, we will focus on further improving our loss function design. For example, can we anchor on the agent policies at different training stages? It will also be interesting to extend the proposed representation learning method in the relaxed setting of IL, like the scene where we can access both the reward and demonstration. 

\section{Reproducibility}
We implement our algorithm according to parameters and details described in Appendix~\ref{sec:algodetails} and Section~\ref{subsec: setup}. We will release our code and data.

\section{Acknowledgement}
This work is supported by the Ministry of Science and Technology of the People´s Republic of China, the 2030 Innovation Megaprojects "Program on New Generation Artificial Intelligence" (Grant No. 2021AAA0150000).  This work is also supported by the National Key R\&D Program of China (2022ZD0161700). We also thanks Guanqi Zhan and Luke Melas-Kyriazi for proof reading.

%% file: appendix.tex
\newpage
\appendix
\onecolumn
\section{Environments}
\label{sec:env}

We use 10 continuous control tasks from the DeepMind control suite~\citep{tassa2018deepmind}. The summary for each task is provided in Table~\ref{tab:tasks}.

\begin{table}[!h]
\centering
\begin{tabular}{lccc}
\hline
Task &$\mathrm{dim}(\gS)$ & $\mathrm{dim}(\gA)$   \\

\hline
Finger Spin   & $6$ & $2$ \\
Hopper Stand  & $14$ & $4$ \\
Pendulum Swingup    & $2$ & $1$ \\
Walker Stand   & $18$ & $6$\\
Walker Walk    & $18$ & $6$ \\
Acrobot Swingup   & $4$ & $1$  \\
Cheetah Run   & $18$ & $6$  \\
Hopper Hop   & $14$ & $4$  \\
Quadruped Run   & $56$ & $12$  \\
Walker Run & $18$ & $6$  \\
\hline
\end{tabular}
\caption{A detailed description of each tasks used in our experiments.}
\label{tab:tasks}
\end{table}

\paragraph{Demonstrations}
For each task, we train expert policies using DrQ-v2~\citep{yarats2021mastering} on the actual environment rewards. We run 3 seeds and pick the seed that achieves the highest return. Then we use this expert policy to collect 10 demonstrations.

\section{Algorithm Details}
\label{sec:algodetails}

\subsection{Implementation}

\paragraph{RL agent}
We use DrQ-v2 as the underlying RL algorithm. DrQ-v2 is an off-policy actor-critic algorithm for continuous control. The core of DrQ-v2 is Deep Deterministic Policy Gradient (DDPG)~\citep{lillicrap2015continuous} augmented with $n$-step returns. 
The critic is trained using clipped double Q-learning~\citep{fujimoto2018addressing} to reduce the overestimation bias in the target value. 
The deterministic actor is trained using deterministic policy gradients (DPG)~\citep{silver2014deterministic}. We also follows the setting of actor's and critic's neural network architectures in state-based DrQ-v2~\cite{yarats2021mastering}.


\paragraph{Contrastive encoder}
The contrastive encoder is implemented as a 4 layer MLP with hidden size [256, 256, 256]. The output dimension is 64. Following the architecture in \cite{yarats2021mastering}, the contrastive encoder, the critic and the actor share the same encoder backbone. This shared encoder is trained with the gradient of the critic alone, which is also following the suggestion of~\cite{kostrikov2020image, yarats2021mastering}. The input of this shared encoder is state $s$ of a transition.

\paragraph{Reward predictor}
Reward of the agent transition is computed according to Equation~\ref{eq:rew}. Thus, we randomly sample one expert transition from the expert replay buffer to compute agent reward. Empirically, we find that using the mean embedding of the expert data yields similar performance.

\paragraph{Gradient penalty}
In order to make the algorithm more stable, we use the gradient penalty technique~\citep{gulrajani2017improved} widely used in Wasserstein-GANs~\citep{arjovsky2017wasserstein}.  We make minor adjustments to accommodate our policy contrastive loss. GAIL-like methods usually constrain the gradient norm of the discriminator’s output with respect to its input. While for PCIL, the contrastive encoder’s output needs one more step. Specifically, the output embeddings are first used to calculate rewards following Equation~\ref{eq:rew}, then we compute and penalize the gradient norm of the rewards. We use $10$ as the weighting for the gradient penalty. 

\subsection{Hyperparameters}
Table \ref{table:hyper} lists the hyperparameters that are used for all baseline methods and our method. Expert data ratio in PCIL means the ratio between expert data and batch size. A ratio of 0.5 means that half of the batch is expert data and the other half is the agent data. The contrastive learning usually needs a temperature scaling after computing the cos-similarity, before computing the exponential. For simplicity, we ignored it in the main text. In the experiment, we follow prior contrastive learning work~\cite{he2020momentum} and use a typical value of $0.07$ for the temperature. 

\begin{table}[h]
\centering
\begin{tabular}{lcc}
\toprule
Methods & Parameter & Value \\
\midrule
all methods         & Replay buffer size  & 500k \\
                    & Agent update frequency & 2 \\
                    & Optimizer   & Adam \\
                    & Learning rate  & 1e-4 \\
                    & Critic soft-update rate & 0.01 \\
                    & Random seed & 1,2,3 \\
                    & RL batch size & 128 \\
                    & Discriminator training batch size & 256 \\
                    & Hidden dim & 256 \\
\midrule
PCIL                & Expert data ratio & 0.5 \\
                    & Contrastive temperature & 0.07 \\
\bottomrule                    
\end{tabular}
\caption{The hyperparameters of baseline methods and our method. }
\label{table:hyper}
\end{table}

\section{Ablation study implementation details}
\label{subsec: ablation detail}
\subsection{TCN representation}
TCN encoder shares the same network architecture as the PCL encoder. During the training process in TCN encoder, for a random sampled anchor, we use data point adjacent to it as positive pair and another random sampled data point as negative pair. We also set contrastive temperature to $0.07$ and batch size to $256$, which is the same as PCL encoder.
\subsection{GAIL-like reward}
GAIL reward predictor can be seen as a simplified version of GAIL discriminator which has only one linear classifier layer. The reward predictor is trained independently to distinguish whether the input is from a expert data or non-expert data with a binary classification loss.

\section{Additional Experimental Results}
\label{sec:more-results}
\begin{figure*}[h]
\centering
\includegraphics[scale=0.55]{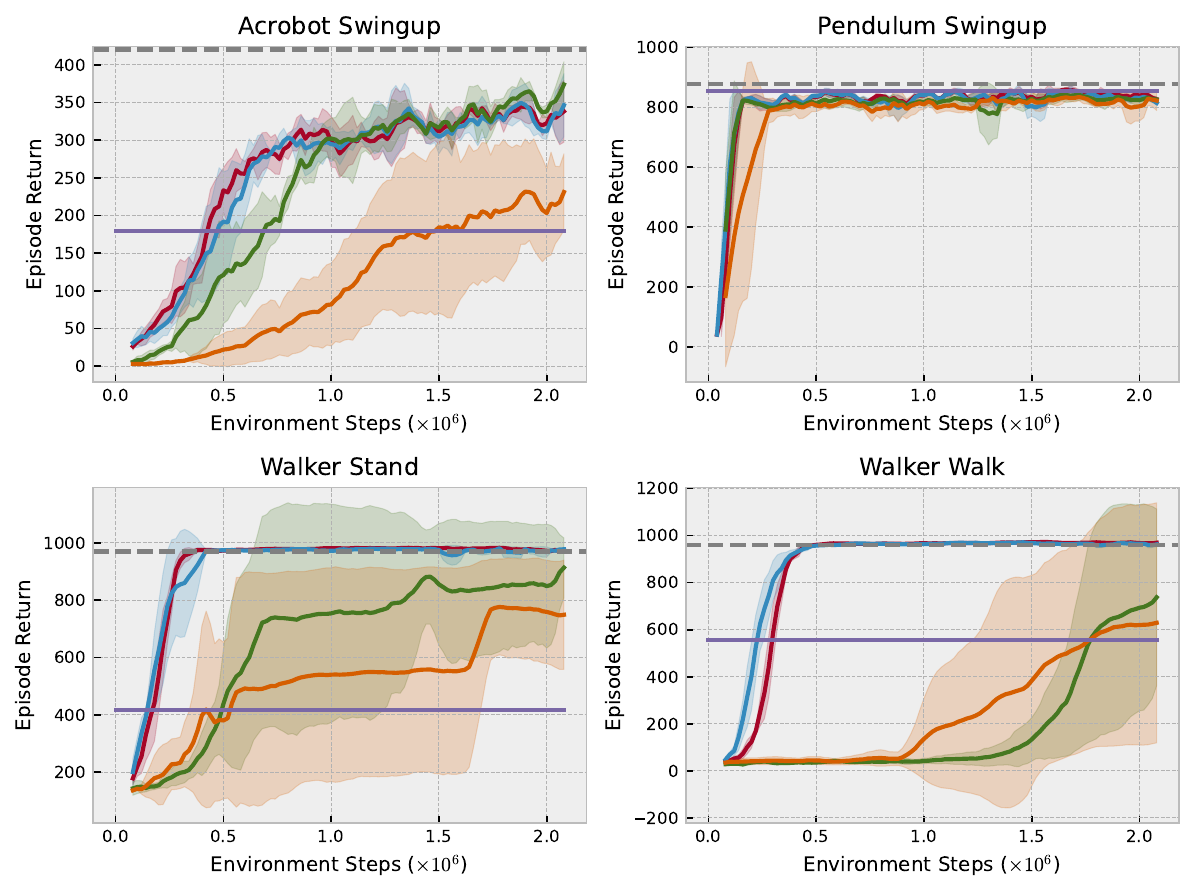}
\\
\cblock{128}{128}{128}\hspace{1mm}Expert\hspace{1.5mm}
\cblock{122}{104}{166}\hspace{1mm}BC\hspace{1.5mm}
\cblock{213}{94}{0}\hspace{1mm}SIL\hspace{1.5mm}
\cblock{70}{120}{33}\hspace{1mm}PWIL\hspace{1.5mm}
\cblock{52}{138}{189}\hspace{1mm}DAC\hspace{1.5mm}
\cblock{166}{6}{40}\hspace{1mm}PCIL (Ours)\hspace{1.5mm}
\caption{Comparisons of algorithms on the other 4 tasks.}
\label{fig:bench-4}
\end{figure*}

Figure~\ref{fig:bench-4} shows the performance of PCIL on the other 4 tasks from the DeepMind Control suite. We notice that the performance of some relatively easy tasks has saturated. All the baselines achieve expert performance on \textit{Pendulum Swingup}. On \textit{Walker Stand} and \textit{Walker Walk}, PCIL is competitive with DAC, which already demonstrates impressive sample efficiency.

\section{Proof}
\label{ap:proof}
At the last of Section 3.4, we have shown that
$$
    \min_\pi\max_{r_\theta} \mathbb{E}_{x\sim \mathcal{D}} r_\theta(x) - \mathbb{E}_{x\sim \rho^\pi} r_\theta(x),
$$
where $r_\theta = (\mathbb{E}_{x'\sim\mathcal{D}}\Phi_\theta(x')dx')^T\Phi_\theta(x)$, and $\mathcal{D}$ is the expert dataset. We may as well regart $\mathcal{D}$ as $\rho^E$ when the dataset size is large. \ \\ \\
Define the inner objective as
$$
D_{cont}(\rho^E||\rho^\pi) = \max_{r_\theta} [\mathbb{E}_{x\sim \rho^E} r_\theta(x) - \mathbb{E}_{x\sim \rho^\pi} r_\theta(x)].
$$
In this part, we prove \ \\ \\
\textbf{Theorem 1}
$$
    0.25 D_{TV}(\rho^E||\rho^\pi)  \leq D_{cont}(\rho^E||\rho^\pi) \leq 2.0 D_{TV}(\rho^E||\rho^\pi).
$$
Here, $D_{TV}(\rho^E||\rho^\pi) = \frac{1}{2}\int_\mathcal{X} |\rho^E(x) - \rho^\pi(x)|dx$ is the total variation divergence. \ \\ \\
\textit{Proof of Theorem 1} \ \\ \\
\textbf{Step 1} We first turn the inner optimization to a simple form. Let $v_\theta = \mathbb{E}_{x'\sim\mathcal{D}}\Phi_\theta(x')$, then we find a corresponding orthonormal matrix $M_\theta$ such that $M_\theta v_\theta  = [\Vert v_\theta\Vert, 0, 0, 0, ..., 0]^T$. As a result, 
$$
r_\theta = (\mathbb{E}_{x'\sim\mathcal{D}}\Phi_\theta(x'))^T\Phi_\theta(x) = (M_\theta\mathbb{E}_{x'\sim\mathcal{D}}\Phi_\theta(x'))^T M_\theta\Phi_\theta(x) = \Vert v_\theta\Vert [M_\theta\Phi_\theta(x)]_1.
$$

If we define $g_\theta = [M_\theta\Phi_\theta(x)]_1: \mathcal{X}\to [-1, 1]$, then the inner objective can be written as:

$ \mathsf{InnerOPT}$
Maximize (w.r.t. $\theta$, $\alpha$)
$$
    \alpha \int_\mathcal{X} g_\theta(x) [\rho^E(x) - \rho^\pi(x)] dx,
$$
subject to 
$$
    \alpha = \left|\int_\mathcal{X} g_\theta(x) \rho^E(x)dx\right|.
$$
Note that the constraint is derived from the following fact:
$$
\alpha = \Vert v_\theta\Vert = \Vert M_\theta v_\theta\Vert = \Vert  \mathbb{E}_{x'\sim\rho^E}M_\theta\Phi_\theta(x')\Vert = \left|\int_\mathcal{X} g_\theta(x) \rho^E(x)dx\right|.
$$
\ \\ \\
\textbf{Step 2} In this step, we show that $0.5 D_{TV}(\rho^E||\rho^\pi)  \leq D_{cont}(\rho^E||\rho^\pi)$. We do this by constructing a $g_\theta$, and evaluate the corresponding objective function. First, let $S = \{x\in\mathcal{X}: \rho^E(x)\geq\rho^\pi (x)\}$. Then, we split the discussion into two cases.
\ \\ \\
\textbf{Case 1} If $\mu^E(S)\geq \mu^E(S^c)$.\ \\ \\
we define 
$$
    g_\theta(x) = \begin{cases}
1 &\text{$x\in S$},\\
-\beta &\text{$x \in S^c$}
\end{cases} .
$$
where $\beta\in [0, \mu^E(S)]$ is a scalar. Then, we know that
$$
    \alpha = |\mu^E(S) - \beta \mu^E(S^c)| = |\mu^E(S) - \beta (1 - \mu^E(S))| = |(1+\beta)\mu^E(S) - \beta| = (1+\beta)\mu^E(S) - \beta.
$$
Meanwhile, we know
$$
\int_\mathcal{X} g_\theta(x) [\rho^E(x) - \rho^\pi(x)] dx = \int_S g_\theta(x) [\rho^E(x) - \rho^\pi(x)] dx + \int_{S^c} g_\theta(x) [\rho^E(x) - \rho^\pi(x)] dx
$$
and
$$
RHS = \int_S |\rho^E(x) - \rho^\pi(x)| dx + \beta\int_{S^c} |\rho^E(x) - \rho^\pi(x)| dx \geq \beta \int_\mathcal{X} |\rho^E(x) - \rho^\pi(x)|dx
$$
Therefore, we have 
\begin{align*}
 D_{cont}(\rho^E||\rho^\pi) &\geq \alpha \int_\mathcal{X} g_\theta(x) [\rho^E(x) - \rho^\pi(x)] dx \geq ((1+\beta)\mu^E(S) - \beta)\beta \int_\mathcal{X}|\rho^E(x) - \rho^\pi(x)|dx \\ &=  2((1+\beta)\mu^E(S)-\beta)\beta D_{TV}(\rho^E||\rho^\pi).
\end{align*}
Since $\mu^E(S)\geq \mu^E(S^c)$ and $ \mu^E(S) + \mu^E(S^c) = 1$, we have $\mu^E(S)\geq 0.5$. Then, 
$$
RHS = 2((1+\beta)\mu^E(S)-\beta)\beta D_{TV}(\rho^E||\rho^\pi) \geq (1-\beta) \beta D_{TV}(\rho^E||\rho^\pi).
$$
In particular, if we pick $\beta = 0.5$, then the right hand side is $0.25D_{TV}(\rho^E||\rho^\pi)$, so in this case,
$$
D_{cont}(\rho^E||\rho^\pi) \geq 0.25D_{TV}(\rho^E||\rho^\pi).
$$
\ \\ \\
\textbf{Case 2} If $\mu^E(S) < \mu^E(S^c)$.
This can be done in a similar way. \ \\ \\
We define 
$$
    g_\theta(x) = \begin{cases}
\beta &\text{$x\in S$},\\
-1 &\text{$x \in S^c$}
\end{cases} .
$$
where $\beta\in [0, 0.5]$ is a scalar. Then, we know that
$$
    \alpha = |\beta\mu^E(S) - \mu^E(S^c)| = |\beta\mu^E(S) - (1 - \mu^E(S))| = |(1+\beta)\mu^E(S) - 1| = 1 - (1+\beta)\mu^E(S).
$$
Meanwhile, we know
$$
\int_\mathcal{X} g_\theta(x) [\rho^E(x) - \rho^\pi(x)] dx = \int_S g_\theta(x) [\rho^E(x) - \rho^\pi(x)] dx + \int_{S^c} g_\theta(x) [\rho^E(x) - \rho^\pi(x)] dx
$$
and
$$
RHS = \beta\int_S |\rho^E(x) - \rho^\pi(x)| dx + \int_{S^c} |\rho^E(x) - \rho^\pi(x)| dx \geq \beta \int_\mathcal{X} |\rho^E(x) - \rho^\pi(x)|dx
$$
Therefore, we have 
\begin{align*}
 D_{cont}(\rho^E||\rho^\pi) &\geq \alpha \int_\mathcal{X} g_\theta(x) [\rho^E(x) - \rho^\pi(x)] dx \geq (1 - (1+\beta)\mu^E(S))\beta \int_\mathcal{X}|\rho^E(x) - \rho^\pi(x)|dx \\ &= 2(1 - (1+\beta)\mu^E(S))\beta D_{TV}(\rho^E||\rho^\pi).
\end{align*}
Since $\mu^E(S^c) > \mu^E(S)$ and $ \mu^E(S) + \mu^E(S^c) = 1$, we have $\mu^E(S) < 0.5$. Then, 
$$
RHS = 2(1 - (1+\beta)\mu^E(S))\beta D_{TV}(\rho^E||\rho^\pi) \geq (1-\beta) \beta D_{TV}(\rho^E||\rho^\pi).
$$
In particular, if we pick $\beta = 0.5$, then the right hand side is $0.25D_{TV}(\rho^E||\rho^\pi)$, so in this case,
$$
D_{cont}(\rho^E||\rho^\pi) \geq 0.25D_{TV}(\rho^E||\rho^\pi).
$$
Putting Case 1 and 2 together, we can conclude that
$$
D_{cont}(\rho^E||\rho^\pi) \geq 0.25D_{TV}(\rho^E||\rho^\pi).
$$
\textbf{Step 3} Finally, we show that 
$$
D_{cont}(\rho^E||\rho^\pi) < 2.0D_{TV}(\rho^E||\rho^\pi).
$$
This is actually quite straightforward. Noting that
$$
    \alpha = \left|\int_\mathcal{X} g_\theta(x) \rho^E(x)dx\right| \leq \int_\mathcal{X} |g_\theta(x) \rho^E(x)| dx = \int_\mathcal{X} |g_\theta(x)| \rho^E(x)dx\leq \int_\mathcal{X} \rho^E(x)dx = 1,
$$
we have 
\begin{align*}
    \alpha \int_\mathcal{X} g_\theta(x) [\rho^E(x) - \rho^\pi(x)] dx &\leq |\alpha| |\int_\mathcal{X} g_\theta(x) [\rho^E(x) - \rho^\pi(x)] dx| \leq |\int_\mathcal{X} g_\theta(x) [\rho^E(x) - \rho^\pi(x)] dx| \\&\leq \int_\mathcal{X} |g_\theta(x)||\rho^E(x) - \rho^\pi(x)| dx \leq \int_\mathcal{X} |\rho^E(x) - \rho^\pi(x)| dx = 2D_{TV}(\rho^E||\rho^\pi).
\end{align*}
This then completes the proof. \ \\ \\